\documentclass[journal]{IEEEtran}
\usepackage{amssymb}
\usepackage{multirow,tabularx}
\usepackage{hhline}
\usepackage{url}
\usepackage{ragged2e} 

\usepackage{cite}

\ifCLASSINFOpdf
   \usepackage[pdftex]{graphicx}
\else
\fi
\usepackage{amsmath}
 \usepackage[caption=false,font=footnotesize]{subfig}

\usepackage{stfloats}
\begin{document}
%
\title{Detection and Localization of Robotic Tools in Robot-Assisted Surgery Videos Using Deep Neural Networks for Region Proposal and Detection}
%
%
%

\author{Duygu~Sarikaya,
        Jason~J.~Corso
        and~Khurshid~A.~Guru
\thanks{D. Sarikaya is a PhD candidate at the Department of Computer Science and Engineering, SUNY Buffalo, NY, 14260-1660 USA e-mail: duygusar@buffalo.edu}
\thanks{J. J. Corso is an Associate Professor at the Department of Electrical Engineering and Computer Science, University of Michigan, Ann Arbor, MI, USA 48109}
\thanks{K. A. Guru is the director of Applied Technology Laboratory for Advanced Surgery (ATLAS), Roswell Park Cancer Institute, Buffalo, NY, USA, 14263}

\thanks{Manuscript received --- 2017.}}

%
%

\markboth{}%
{Shell \MakeLowercase{\textit{et al.}}: Bare Demo of IEEEtran.cls for IEEE Journals}
%



\maketitle

\begin{abstract}
Video understanding of robot-assisted surgery (RAS) videos is an active research area. Modeling the gestures and skill level of surgeons presents an interesting problem. The insights drawn may be applied in effective skill acquisition, objective skill assessment, real-time feedback, and human-robot collaborative surgeries. We propose a solution to the tool detection and localization open problem in RAS video understanding, using a strictly computer vision approach and the recent advances of deep learning. We propose an architecture using multimodal convolutional neural networks for fast detection and localization of tools in RAS videos. To our knowledge, this approach will be the first to incorporate deep neural networks for tool detection and localization in RAS videos. Our architecture applies a Region Proposal Network (RPN), and a multi-modal two stream convolutional network for object detection, to jointly predict objectness and localization on a fusion of image and temporal motion cues. Our results with an Average Precision (AP) of $91\%$ and a mean computation time of $0.1$ seconds per test frame detection indicate that our study is superior to conventionally used methods for medical imaging while also emphasizing the benefits of using RPN for precision and efficiency. We also introduce a new dataset, ATLAS Dione, for RAS video understanding. Our dataset provides video data of ten surgeons from Roswell Park Cancer Institute (RPCI) (Buffalo, NY) performing six different surgical tasks on the daVinci Surgical System (dVSS\textsuperscript{\textregistered}) with annotations of robotic tools per frame. 
\end{abstract}

\begin{IEEEkeywords}
Object detection, Multi-layer neural network, Image classification, Laparoscopes, Telerobotics 
\end{IEEEkeywords}

%
\IEEEpeerreviewmaketitle

\section{Introduction}
\begin{figure*}[!t]
\centering
\includegraphics[width=3.6in]{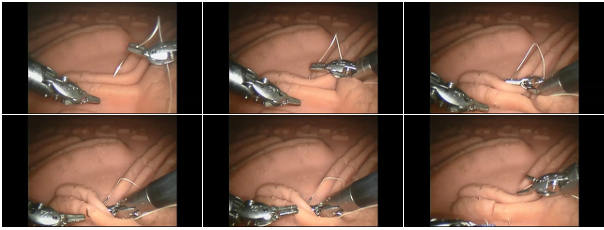}{\includegraphics[width = 3.6in]{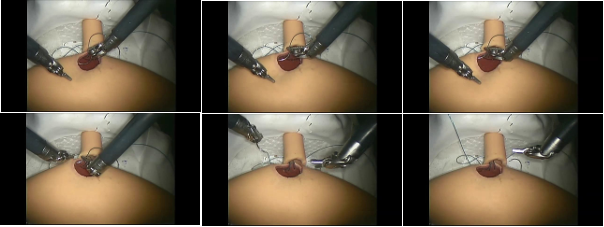}}
\caption{ Sample frames of the video data recorded during surgical training tasks on the daVinci Surgical System (dVSS\textsuperscript{\textregistered}).} \label{fig:videodata}
\end{figure*}

%
%
%
%
\IEEEPARstart{R}{obot-assisted} surgery (RAS) is the latest form of development in today's minimally invasive surgical technology. The robotic tools help the surgeons complete complex motion tasks during procedures with ease by translating the surgeons' real-time hand movements and force on the tissue into small scale ones. Despite its advances in minimally invasive surgery,  the steep learning curve of the robot-assisted surgery devices remains a disadvantage \cite{meadows}. Translation of the surgeons' movements via the robotic device is challenged by a loss of  haptic sensation. Surgeons usually feel comfortable with the procedures only after they have completed procedures on $12$-$18$ patients \cite{meadows}. The traditional mode of surgical training (apprenticeship) fails to answer the needs of today's RAS training. The conventional approach relies heavily on observational learning and operative practice \cite{krovi}. Moreover, the evaluation of the training also relies on the subjective observance of an experienced surgeon, ideally measured by a senior surgeon with a scoring system \cite{race}. The need for universally-accepted and validated metrics and quantitative skill assessment via automation is addressed in the community \cite{krovi}. Early identification of technical competence in surgical skills is expected to help tailor training to personalized needs of surgeons in training \cite{race,krovi}. Moreover, we believe that  automated feedback or human-robot collaborative surgeries could greatly benefit novice surgeons and their patients. As such, we have investigated the development of such systems with video understanding via computer vision using the video data recorded during surgical tasks on the daVinci Surgical System (dVSS\textsuperscript{\textregistered}). Figure \ref{fig:videodata} shows sample frames from recorded video data of surgeons performing on training sets.

\begin{figure}[!t]
\raggedleft
{\includegraphics[width = 3.6in]{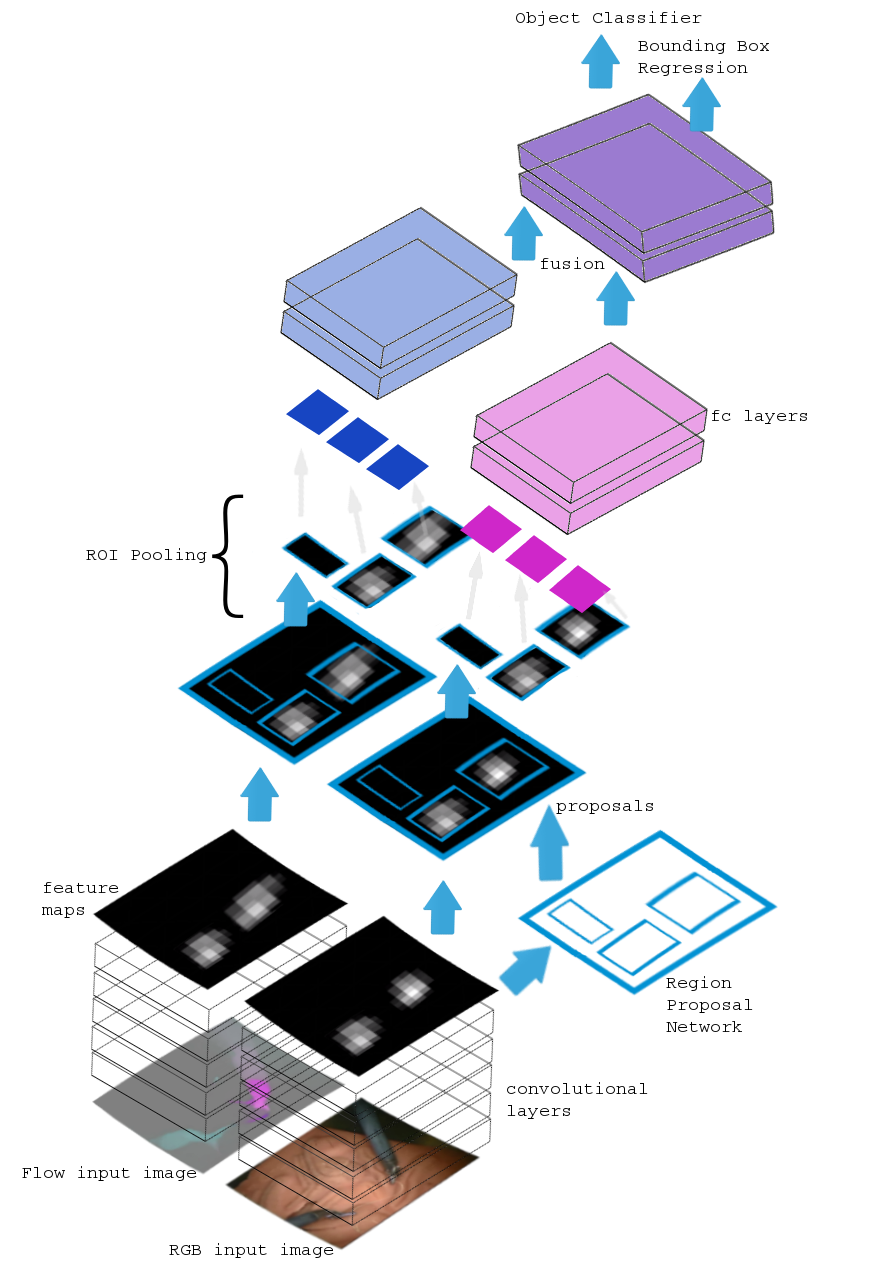}}
\caption{We propose an end-to-end deep learning approach for tool detection and localization in RAS videos. Our architecture has two separate CNN processing streams on two modalities: the RGB video frame and the RGB representation of the optical flow information of the same frame.  We convolve the two separate input modalities and get their convolutional feature maps. Using the RGB image convolutional features, we train a Region Proposal Network (RPN) to generate object proposals. We use the region proposals and the feature maps as input to our object classifier network. The last layer features of these streams are later fused together before classifier. } 
\label{fig:overview}
\end{figure}

We approach the problem of video understanding of RAS videos as modeling the motions of surgeons. Modeling gestures and skills depends highly on the motion information and the precise trajectories of the motion. We argue that, detecting the tools and capturing their motion in a precise manner is an important step in RAS video understanding.  The first step for tracking the tools in the video is to robustly detect the presence of a tool and then localize it in the image. Tool detection and localization is hence the focus of our study.

Advances in object detection had plateaued until the recent reintroduction of deep neural networks into the computer vision community with large-scale data object classification tasks. The deep neural network trained via back-propagation through layers of convolutional filters by LeCun et al. \cite{lecun89,lecun98} had had an outstanding performance on large amounts of training data. On large-scale recognition challenges like ImageNet \cite{imageNet}, similar approaches have now proven that deep neural networks could be used with object recognition and classification tasks \cite{imageNet,Krizhevsky}. However, in the medical field these advances are yet unused, to the best of our knowledge; so, we propose a novel approach of using deep convolutional neural networks (CNN) for fast detection and localization for video understanding of RAS videos. No study we are aware of has addressed the use of CNN in tool detection and localization in RAS video understanding and our study will be the first. 

In this paper, we address the problem of object detection and localization, specifically the surgical robotic tool, in RAS videos as a deep learning problem. We propose complete solutions for detection and localization using CNNs. We apply Region Proposal Networks (RPN) \cite{fasterRcnn} jointly with a multimodal object detection network for localization; as such, we simultaneously predict object region proposals and objectness scores. Figure \ref{fig:overview} shows an overview of our system. Our results indicate that our study is superior to conventionally used methods for medical imaging with an Average Precision (AP) of $91\%$ and a mean computation time of $0.1$ seconds per test frame.

\section{Related Work}
Many early approaches to detecting and tracking robotic tools in surgery videos use markers and landmarks to reduce the problem to a simple computer vision problem of color segmentation or thresholding \cite{Groeger,Wei}. Using color markers or color coding the tools are examples of this approach. Another example of a marker is a laser-pointing instrument holder used to project laser spots on the scene \cite{Krupa}. The works of Zhang et al. \cite{payandeh} and Zhao et al. \cite{zhao} introduce a barcode marker. However, these methods require additional manufacturing and raise concerns on bio-compatibility. Moreover, having to use additional instruments in minimally-invasive surgical settings could be challenging \cite{pote}. 

More recent approaches focus on tracking the tool via a per video initialization and/or using additional modalities such as kinematic data.  Du et al. develop a 2D tracker based on a Generalized Hough Transform using SIFT features and use this to initialize a 3D tracker at each frame to recover instrument pose \cite{du,ballard,sift}. This method assumes that they have the 3D pose of the instrument in the first frame and they use this information to initialize a 2D bounding box which is then used as a reference point for tracking. Allan et al. uses Random Forests to fuse region based constraints based on multi-label probabilistic region classifications with low level optical flow information \cite{allan}. Instead of an offline learning approach, they train their Random Forest using a manual segmentation of a single frame with a tool positioned in front of the scene. This approach poses a drawback as it does not allow their system to operate on different surgical setups without re-training. Capturing kinematic data of the robotic console as an additional feature to help with tool detection and localization could be an alternative, however, it requires additional instruments,  recording, and preprocessing in order to be used as motion information. Recent approaches such as the work presented in Reiter et al.'s \cite{reiter}, creates templates using different kinematic configurations of the virtual renderings of a CAD model of the tool. They further refine their configuration estimation by matching gradient orientation templates to the real image.

\begin{figure}[!t]
\centering
{\includegraphics[width = 3.5in]{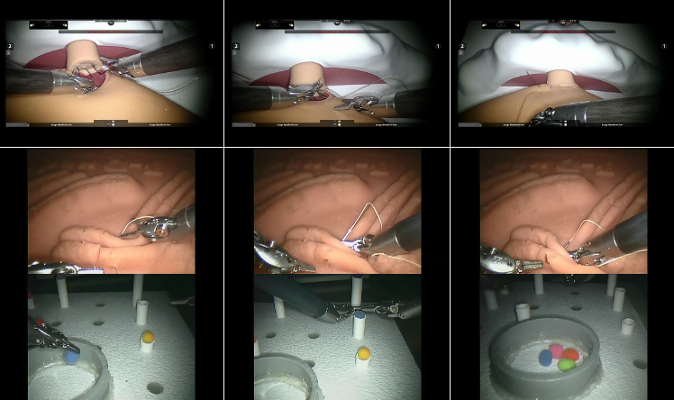}}
\caption{The tools in RAS videos vary in pose, articulation and parts of the tool might be occluded or missing in the frame.} 
\label{fig:occlusion}
\end{figure}

There are strictly vision based approaches as well. In their work, Sznitman et al. \cite{sznitman} learn a multiclass classifier based on tool-parts detectors. During training they use image windows and evaluate the image features that compute the proportions of edges in different locations and orientations in relation to that window. At test time, to localize the tool, they evaluate the classifier in a sliding window fashion; at each image location and at multiple image scales. Although they use an early stopping algorithm to reduce the time, we believe that it is a priority to offer a solution that is more effective and efficient. One of the common computer vision approaches is using shape matching. The challenge of detection by fitting shape models is that the objects in the RAS videos in study show a high-degree of variation in shape, pose, and articulation. The parts of the objects might even be occluded or missing in the frame as seen in Figure \ref{fig:occlusion}. So the approaches that follow a rigid appearance and global shape model are usually not sufficient.  In contrast, the Deformable Parts Model (DPM) by Felzenszwalb et al. \cite{dpm,objdetdpm} consists of a star-structured pictorial model linking the root of an object to its parts using deformable springs. This model has proven to be successful in object detection as it captures the occlusions and articulations. One of the recent,  surgical tool detection applications of DPM \cite{dpm,objdetdpm} is the product of experts tool detector by Kumar et al. \cite{pote}. We compare our results to DPM \cite{dpm,objdetdpm} as it is often successfully used in recent successful applications in similar domains. 

In this paper, we propose a novel end-to-end approach of using deep convolutional neural networks (CNN) for fast detection and localization for video understanding of RAS videos. We apply Region Proposal Networks (RPN) \cite{fasterRcnn} jointly with a multimodal object detection network for localization.  Our architecture, based on the work of Zeiler et al. \cite{zeiler} and developed to support multimodality, starts with two separate CNN processing streams on two modalities: the RGB video frame and the RGB representation of the optical flow information of the same frame as temporal motion cues. RPN is a deep, fully convolutional network that outputs a set of region proposals based on the convolutional feature maps of the RGB image inputs.  We adopt Fast R-CNN  \cite{fastRcnn} for the object detection task, we use the region proposal boxes of RPN with the convolutional features as input for the detection network streams on both modalities. The last layer features of these streams are fused together. 

Our study differs from the former approaches as we focus on proposing a solution using strictly computer vision instead of using additional modalities that requires additional equipment to capture; such as kinematic data. With our study, we propose to make use of recent advances of Convolutional Neural Networks with the hope that it will serve as a benchmark for tool tracking to move towards deep learning. We also prioritize an efficient and effective way of localizing the tools in the images, our study demonstrates the invaluable contribution of RPNs to the problem of tool detection and localization in RAS videos. With our new architecture that incorporates two modalities, we make use of not only the visual object features but also the temporal motion cues. This approach helps us improve the detection of the tools by decreasing false positives. Our experiments show that our method is able to detect and localize the tools fast with superior accuracy. Our proposed method does not require initialization, so it can be used on new video data without re-training. It could be used to automatically initialize tracking algorithms in RAS videos. 

\begin{table*}
\centering
\caption{Properties of the Dataset}
\label{table:data}
\begin{tabular}{lllll}
\textit{\textbf{Main Dataset}}                                          &                                                  &                           &                                    &                                \\
\textit{\textbf{Skill Level}}                                                       & \textit{\textbf{Task}}                           & \textit{\textbf{Subtask}} & \textit{\textbf{Number of Videos}} &\textit{\textbf{ Number of Frames}}               \\ \hline
\multirow{6}{*}{Basic Skills}                                                       & \multirow{2}{*}{Ball Placement Task}             & Using 1 Arm               & 7 videos                           &                                \\
                                                                                    &                                                  & Using 2 Arms              & 7 videos                           &                                \\
                                                                                    & \multirow{2}{*}{The Ring Peg Transfer Task}      & Using 1 Arm               & 7 videos                           &                                \\
                                                                                    &                                                  & Using 2 Arms              & 7 videos                           &                                \\
                                                                                    & \multirow{2}{*}{Suture Pass Task}                & Put Through               & 7 videos                           &                                \\
                                                                                    &                                                  & Pull Through              & 7 videos                           &                                \\
\multirow{3}{*}{Intermediate Skills}                                                & \multirow{3}{*}{Suture and Knot Tie Task}        & Suture Pick Up            & 7 videos                           &                                \\
                                                                                    &                                                  & Suture Pull Through       & 7 videos                           &                                \\
                                                                                    &                                                  & Suture Tie                & 7 videos                           &                                \\
\multirow{3}{*}{Advanced Skills}                                                    & \multirow{3}{*}{Urethrovesical Anastomosis(UVA)} & UVA Pick Up               & 7 videos                           &                                \\
                                                                                    &                                                  & UVA Pull Through          & 7 videos                           &                                \\
                                                                                    &                                                  & UVA Tie                   & 7 videos                           &                                \\
\textit{\textbf{Total}}                                                             &                                                  &                           & \textit{\textbf{84 videos}}        & \textit{\textbf{18782 frames}} \\ \hline
\begin{tabular}[c]{@{}l@{}}Additional Test Set \\ of compiled subtasks\end{tabular} & \multicolumn{2}{l}{}                                                         & 15 videos                          &    \textit{{3685 frames}}                              \\
\textit{\textbf{Total}}                                                             & \multicolumn{2}{l}{\textit{\textbf{}}}                                       & \textit{\textbf{99 videos}}        &     \textit{\textbf{22467 frames}}                             \\ \hline
\end{tabular}
\end{table*}

\section{Dataset}

RAS video understanding may be particularly challenging as the view of the operating site is limited and recorded via endoscopic cameras. Tracking the free movement of the surgeons in unconstrained scenarios requires camera movement and zoom. The tools in RAS videos vary in pose, articulation and their parts might be occluded or missing in the frame (refer to Figure \ref{fig:occlusion} for visual examples). There are frequently other objects in use and in motion such as the needle, suture, clamps or other objects used in training. The tissue that the surgeon operates on might move or deform, show variation in shape and occlude the tool.

We have noticed that the RAS datasets for public use are limited.  The only comprehensive RAS datasets we know of that is open for public use are JIGSAWS by Gao et al. \cite{jigsaws} and the very recently released m2cai16-workflow and m2cai16-tool datasets \cite{m2cai}. Neither of these datasets provide tool annotations. JIGSAWS is quite restricted as it does not include artifacts, camera movement and zoom, or a wide range of free movement. For this reason, we have built a new, more challenging dataset; ATLAS Dione, that leverages the data gathered for earlier works of Guru et al. \cite{race} and Stegemann et al. \cite{fsrs}. We have used the video data and prepared manual annotations for RAS video understanding problems. This new video dataset is a core contribution of our work and we will release it for tool detection and localization purposes upon publication. Despite being a phantom setting, our dataset is quite challenging as it has camera movement and zoom, free movement of surgeons, a wider range of expertise levels, background objects with high deformation, and annotations include tools with occlusion, change in pose and articulation or when they are only partially visible in the scene.

ATLAS Dione provides video data of ten subjects performing six different surgical tasks on the dVSS\textsuperscript{\textregistered}. The ten surgeons who participated in this IRB-approved study (I 228012) work at Roswell Park Cancer Institute (RPCI) (Buffalo, NY). The difficulty and complexity of the tasks vary. These tasks include basic RAS skill tasks which are part of the Fundamental Skills of Robotic Surgery (FSRS) curriculum \cite{fsrs} and also the prodecure-specific skills required for the Robotic Anastomosis Competency Evaluation (RACE) \cite{race}. The subjects, surgeons from RPCI, are classified by different expertise levels of beginner (BG), combined competent and proficient (CPG), and expert (EG) groups based on the Dreyfuss model \cite{dreyfus}. For tool detection and localization purposes, we manually label both the left and right tools in use by providing exact bounding boxes of tool locations. In addition to these, following our future works, expertise levels of the subjects, task videos with the beginning and ending timestamps for each subtask will also be released for surgical activity recognition research. 

\begin{figure}[!t]
\centering
\subfloat[Basic Skills: Ball Placement Task, Suture Pass Task and Ring Peg Transfer Task.]{\includegraphics[width = 3.5in]{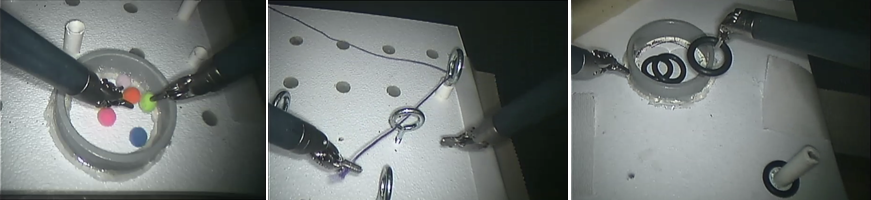}}\\
\subfloat[Intermediate Skills:  Placement of Simple Suture with Knot Tying. Advanced Skills: Performance of Urethrovesical Anastomosis (UVA) on an Inanimate Model.]{\includegraphics[width = 3.5in]{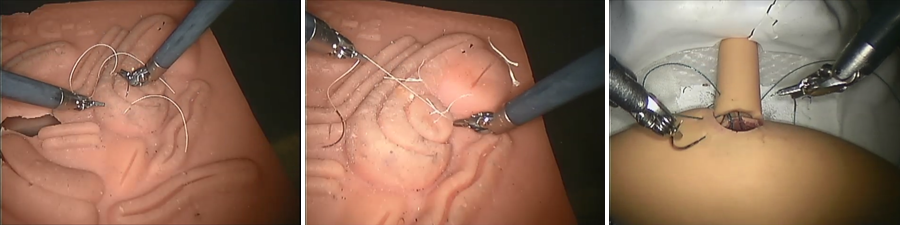}}
\caption{Sample frames of tasks under different robotic skill groups based on the difficulty and complexity.}
 \label{fig:skill}
\end{figure}

\justify
The robotic skill groups based on the difficulty and complexity of the tasks are as follows (Refer to Figure \ref{fig:skill} for sample frames of these skill groups):

\centering
\begin{enumerate}
  \item Basic Skills: Ball Placement Task, Suture Pass Task and Ring Peg Transfer Task.
  \item Intermediate Skills: Placement of Simple Suture with Knot Tying.
  \item Advanced Skills: Performance of Urethrovesical Anastomosis (UVA) on an Inanimate Model.
\end{enumerate}

Please refer to Table \ref{table:data} to see properties of the dataset.

\justify

\subsection{Tool Annotation}

For the tool detection and localization, we annotate bounding boxes for both left and right tools seen in the videos. Bounding boxes are manually annotated with the guidance of an expert RAS surgeon. For efficient annotation, we have customized the tools provided by Caltech Pedestrian Detection Benchmark by Doll\'ar et al.  \cite{caltech} to our problem and used it to annotate each frame in the video clips. 

We provide each frame (each of size 854$x$480 pixels) of the RAS videos in JPEG format. The annotations are provided in xml templates in VOC format \cite{voc}. 

\subsection{Subject Demographics}

We recorded 10 surgeons  from RPCI performing the given tasks on (dVSS\textsuperscript{\textregistered}). Out of these 10 subjects,  2 of them are residents, 3 are fellows and 5 are practicing robotic surgeons. While 3 of them have more than 10 years of experience, 2 of them have between 2 to 5 years of experience and the remaining 5 subjects are still in training.  Two of our subjects have performed over 500 robot-assisted procedures. The ten subjects are  assigned to beginner (BG), combined competent and proficient (CPG), and expert (EG) groups based on the Dreyfuss model \cite{dreyfus}. 

The dataset with tool annotations is available for download to encourage further research in RAS video understanding.

\section{Method}\label{sec:methods}

We propose an end-to-end deep learning approach for tool detection and localization in RAS videos. Our architecture, based on the work of Zeiler et al. \cite{zeiler}, has two separate CNN processing streams on two modalities: the RGB video frame and the RGB representation of the optical flow \cite{flow} information of the same frame.  The last layer features of these streams are later fused together. We first convolve the two separate input modalities and get their convolutional feature maps. Using the RGB image convolutional features, we train a Region Proposal Network (RPN) \cite{fasterRcnn} to generate object proposals. Figure \ref{fig:overview} shows an overview of our system.

\begin{figure}[!t]
\centering
{\includegraphics[width = 3.75in]{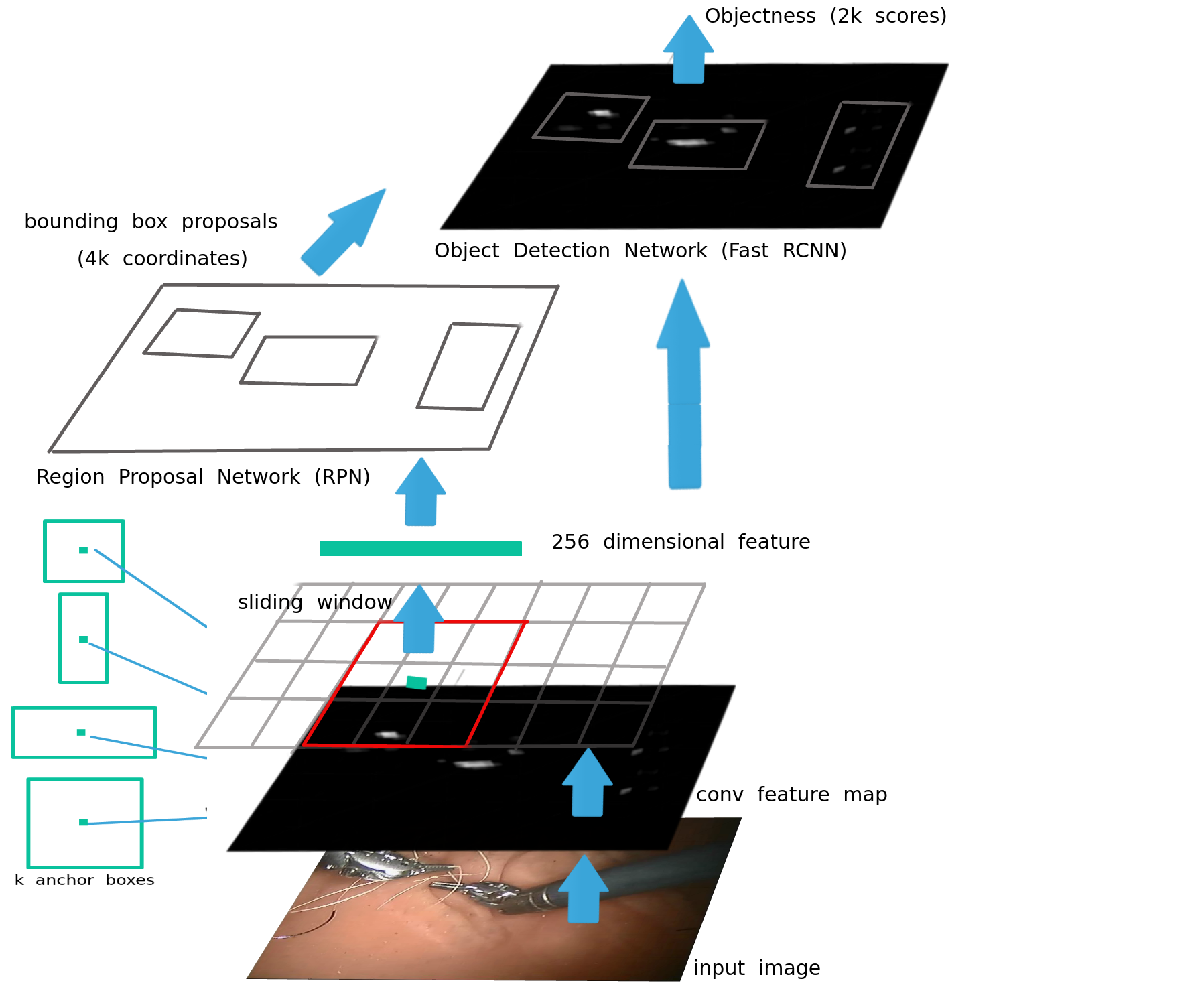}}
\caption{We use a Region Proposal Network that proposes regions, which are used by a detector network. We slide a small network over the convolutional feature map. Each spatial window over the feature map is mapped to a 256 dimensional feature which is fed into the system. RPN outputs object region proposals, each with an objectness score on whether the region is of a tool or not. (Please refer to \ref{sec:methods} for details of our Method.)} 
\label{fig:steps}
\end{figure}

Figure \ref{fig:steps} provides a visual overview of the region proposal network. RPN uses the convolutional feature maps of the RGB input for generating region proposals. Each of these proposals have an objectness score.  In our study, we use an architecture based on the one proposed by Zeiler et al. \cite{zeiler} with five convolutional layers followed by fully connected layers. As proposed by the work of Girschick et al. \cite{fasterRcnn}, we slide a network over the convolutional feature map of the $conv_{5}$ layer in a sliding-window fashion. This network is fully connected to a spatial window of the convolutional feature map with an $3 \times 3$ convolutional layer. Then each $3 \times 3$ window is mapped to a lower-dimensional, fixed $256$ dimensional feature vector. This feature vector is then used as input for a box regression layer and a box classification layer. The regression layer outputs $4k$ values: the bounding box coordinates for each of the $k$ region proposals. The classification layer outputs $2k$ scores that estimate probability of each of the $k$ regions being of object class or not. Region proposals are relative reference boxes to anchors centered at each sliding window. Each anchor is related with a scale of size $128$, $256$, and $512$ pixels and aspect ratios of $1:1$, $1:2$, and $2:1$ resulting in $9$ anchors at each sliding window and $WHk$ translation invariant anchors in total where the convolutional feature map size is  $W  \times H$. At the training step, each anchor is given a binary class label according to Intersection-over-Union (IoU) overlap with a ground-truth box. If the IoU is higher than $0.7$, the anchor is given a positive object class label whereas if the IoU is smaller than $0.3$ the anchor is given a negative label. The remaining anchors are considered neutral and are not used for training purposes.  

We use these region proposal boxes from the RPN and the convolutional features of both modalities, as input to the ROI pooling layer, as introduced by Girschick et al. \cite{fastRcnn}, for each stream. After the last fully connected layers, the features of both streams are concatenated and fused together in a fully connected layer before the loss layers. We train region proposal network and the multimodal object detection network jointly,  with reference to the approximate joint training; that is, we ignore the derivate with respect to the proposal boxes coordinates as explained by Girschick et al. \cite{fasterRcnn}. We also initiate our region proposal network with the RGB modal image only, while we convolve the two modality pair of images and then fuse their high level features for the detection network. We think of the optical flow as a function on the input of RGB image features. Our experiments show that this approach is able to produce competitive results with a modest training time.

\subsection{Loss function for Learning}

We use a multitask objective function \cite{baxter,caruana} to enable learning parameters over our full network concurrently.

In our work, as proposed in Faster R-CNN \cite{fasterRcnn}, we minimize an objective function following the multi-task loss:

\begin{equation}\label{E1}
\resizebox{.445 \textwidth}{!} {
$
L(\{p_{i}\},\{t_{i}\})=\frac{1}{N_{cls}}\sum_{i}L_{cls}(p_{i},p_{i}^{*})+\lambda\frac{1}{N_{reg}}\sum_{i}p_{i}^{*}L_{reg}(t_{i},t_{i}^{*})
$
}
\end{equation}

Recall that the regression layer outputs $4k$ values; the bounding box coordinates for each of the $k$ region proposals while the classification layer outputs $2k$ scores that estimate probability of each of the $k$ region proposals being of object class or not. The classification layer outputs  a discrete probability $\{p_{i}\}$ $p = (p_{0}, . . . , p_{K})$, over $K + 1 (background/non-object)$ categories and the regression layer outputs $\{t_{i}\}$ bounding-box regression offsets; a predicted tuple $t^{u}=(t^{u}_{x},t^{u}_{y},t^{u}_{w},t^{u}_{h})$ for class $u$.  In Equation (\ref{E1}), $i$ is the index of an anchor and $p_{i}$ is the predicted probability of anchor $i$ being an object. According to the IoU overlap, $p^{*}_{i}$ , the ground truth label, is 1 if the anchor is positive, and is 0 if the anchor is negative. $t_{i}$ is the offset of the predicted bounding box where $t^{*}_{i}$ is the offset of the ground-truth box associated with a positive anchor.  The classification loss $L_{cls}$ is log loss over classes for whether it is an object or not:

\begin{equation}
L_{cls}(p, u) = - \log p_{u} 
\end{equation}  
for true class $u$.

For the regression loss ( $L_{reg}$), we use $L_{reg}(t_{i},t^{*}_{i})$ =  $R(t_{i}-t^{*}_{i})$, where R is the robust loss (smooth L1) function shown below:

\begin{equation}
smooth_{L1}(x)  =\left\{
  \begin{array}{@{}ll@{}}
    0.5 x^{2} & \text{if}\ |x| < 1 \\
    |x| - 0.5 & \text{otherwise}
  \end{array}\right.
\end{equation}

Regression loss is activated only for positive anchors and does not contribute otherwise. The two terms are normalized with $N_{cls}$ and $N_{reg}$ and a balancing weight $\lambda$, set to 10, that controls the balance between the two task losses. $cls$ is normalized by the mini-batch size (i.e., $N_{cls}$ = 256) and the $reg$ is normalized by the number of anchor locations. For a convolutional feature map of a size $W \times H$, there are $WHk$ anchors in total.

We use the bounding-box regression to predict more precise boundaries and to improve localization. At each object proposal bounding box, predictions are refined with the help of the anchors and regression. The features are of a fixed size.  However, a set of $k$ bounding-box regressors for each proposed box, called anchors, are learned. Each of these anchors are responsible for a scale and an aspect ratio (Figure \ref{fig:steps}).

A transformation that maps an anchor of a proposed box $P$ to a nearby ground-truth box $G$ is learned using the regression suggested by Girschick et al. \cite{Rcnn}. $P^{k}=(P^{i}_{x},P^{i}_{y},P^{i}_{w},P^{i}_{h})$ is defined to be the  pixel coordinates of the center of an anchor box of a proposal $P$ with $P$'s width and height. Similiarly, the ground truth pair is $G=(G_{x},G_{y},G_{w},G_{h})$. The transformation between the pairs of $P$ and the nearest ground-truth box $G$ is parameterized as a scale-invariant translation of the center of $P$’s bounding box $d_{x}(P)$,$d_{y}(P)$ and log-space translations of the width and height of $P$’s bounding box.

We apply the transformations below  from $P$ into $G$:

\begin{align*}
\hat{G}_{x}  = P_{w}d_{x}(P) + P_{x} \\
\hat{G}_{y}  = P_{h}d_{y}(P) + P_{y} \\
\hat{G}_{w} = P_{w} \exp(d_{w}(P)) \\
\hat{G}_{h}  = P_{h} \exp(d_{h}(P))
\end{align*}

Each function $d_{\star}(P)$ is a linear function of the pool5
features of  $P$. Assuming  $d_{\star}(P)=w^{T}\Phi_{5}(P)$   

We learn $w_{\star}$ by ridge regression:

The regression targets $t_{\star}$ for the training pair ($P$, $G$) are then defined as:

\begin{align*}
t_{x} = (G_{x} − P_{x})/P_{w} \\
t_{y} = (G_{y} − P_{y})/P_{h} \\
t_{w} = \log(G_{w}/P_{w}) \\
t_{h} = \log(G_{h}/P_{h})
\end{align*}
which we solve as a standard regularized least squares problem.

\subsection{Optimization}

The RPN is a fully-convolutional network trained end-to-end with back-propogation and stochastic gradient descent (SGD). To train this network efficiently, we first sample $N$ images and then sample $R/N$ anchors from each image pair. We also randomly sample a top-ranking fixed number of anchors in each modality image to compute the loss function of a mini-batch and keep a ratio of $1:1$ of positive and negative anchors. 

During the optimization we initialize the new layers by weights from  a zero-mean Gaussian distribution with standard deviation $0.01$, while using the pre-trained ImageNet \cite{imageNet} model to initialize the rest.  Transferring weights from pre-trained Imagenet model has helped greatly with initialization of RPN networks and also with the RGB image convolutional features. It has also shown great benefits by addressing the problem of overfitting and has decreased the fluctuations while our model converged. Unfortunately, we couldn't find a suitable dataset model to train the newly introduced layers for the optical flow modality, we have initialized these layers by weights from a zero-mean Gaussian distribution. This has led to some fluctuations while the model converged, however, in the end, has shown improvement with accuracy. 

 We set the learning rate to $0.001$, with momentum of $0.9$ and a weight decay of $0.0005$. We set the number of iterations to 70$k$, we have decided on this number to be optimal after experimenting with a range of 30$k$ to 120$k$ iterations. 70$k$ has shown comparable results to 120$k$ while taking much less time for training.  

\subsection{Detection at test time}

During testing we apply the fully-convolutional RPN to the entire image. To reduce the redundancy of overlapping proposals, we use non-maximum suppression (NMS) with the threshold of $0.7$ on the proposals based on their objectness scores. Then we use the top ranking fixed number of proposals for detection at test.

\section{Experiments and Evaluation}\label{Experiments}

We evaluate our architecture on the ATLAS Dione dataset using all the 99 videos for either training or testing. In order to experiment how stable the average precision results are, we do a ten fold experiment. We split 90 of the videos for training and the rest 9 for testing for each experiment setup. The videos are randomized for each experiment and we never use the frames extracted from the same video for both training and testing for the reason they might be too similar. Our dataset has only one class object; that is, the robotic tool and the background class.  We use a setting of multiple $2.62$GHz CPU processors and Geforce GTX$1080$ with computation capability of  $6.1$, which have allowed us to run our experiments faster with less memory consumption with cuDNN library. We use PASCAL VOC evaluation \cite{voc} to evaluate the accuracy of our detections.

\begin{table*}
\centering
\caption{Experiments and Evaluation}
\label{table:experiments}
\begin{tabular}{lllll}
\textit{\textbf{Method}}                                                      							 & \textit{\textbf{Mean Average Precision}}                           & \textit{\textbf{Detection Time (per frame)}}                \\ \hline
                                                                                                                  					
\multirow{3}{*}{RPN+Fast R-CNN Detection, Multimodal}                                                         & \multirow{3}{*}{ $91\%$ ($90.65\%$)}         &  \multirow{3}{*}{ $0.103$ seconds }                \\        
																		 &                                                                                     &   \multirow{3}{*}{      + optical flow computation (a few seconds \cite{flow})  }                                     \\
																	          &									     &																					\\
\multirow{3}{*}{RPN+Fast R-CNN Detection (Faster R-CNN)}                                                    & \multirow{3}{*}{ $90\%$ ($0.9039\%$)}       &  \multirow{3}{*}{ $0.059$ seconds}                                    \\         
																		 &                                                                                     &                                                           \\      
 \multirow{3}{*}{Edge Boxes+Fast R-CNN Detection (Fast R-CNN)}                                        & \multirow{3}{*}{ $20\%$} 				     &  \multirow{3}{*}{ $0.134$ seconds for detection }                                    \\ 
																		 &                                                                                     &    \multirow{3}{*}{         + 2 seconds for region proposal     }                                          \\         
  																		&									     &																					\\    
\multirow{3}{*}{Deformable Parts Model (DPM)}                                                   			 & \multirow{3}{*}{$76\%$ ($83\%$ with bounding box regression)} 				     &  \multirow{3}{*}{ $2.3$  seconds}                              \\                               
									  									 &                                                                                     &                                                           \\                  

\end{tabular}
\end{table*}

\begin{figure}[!t] 
\centering
\subfloat[Sample detection results of regular scenes with their scores. ]{\includegraphics[width = 3.5in]{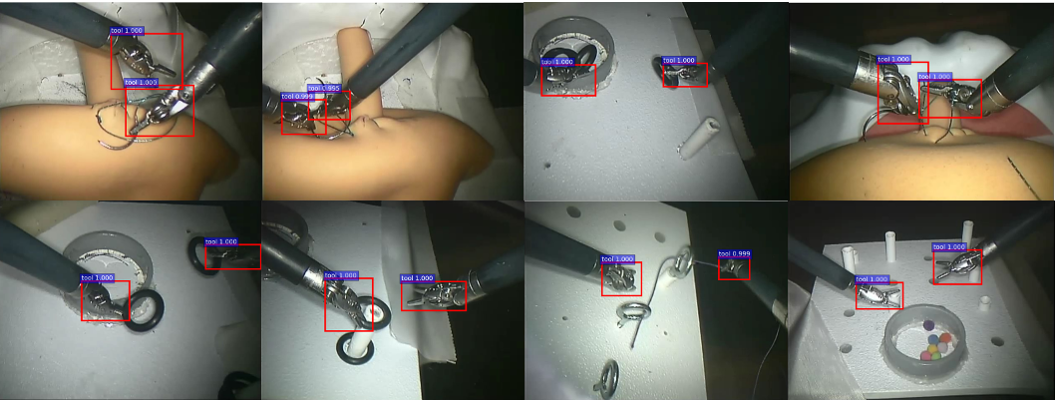}}\\ 
\subfloat[Sample detection results of challenging scenes with their scores. Here we can see confident detections with precise boundaries even when the tool is occluded or partially out of screen.]  {\includegraphics[width = 3.5in]{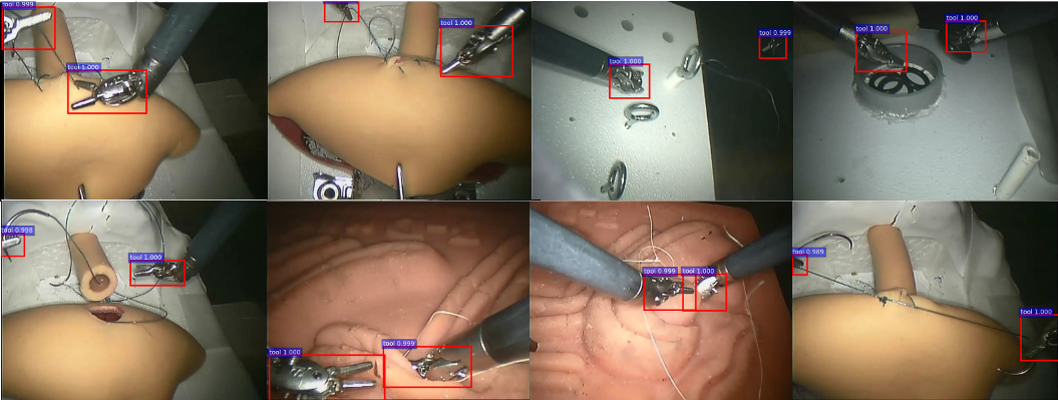}}
\caption{Sample detection results.}
\label{fig:results}
\end{figure}

Our experimental results reach an Average Precision (AP) of $91\%$ ( $90.65\%$). It takes $7.22$ hours to train a model with $70k$ iterations and a mean computation time of $0.103$ seconds to detect the tools in each test frame, given that we are working with a set of already computed optical flow images. We compare our experimental results with the original architecture proposed by Girschick et al. \cite{fasterRcnn}, Fast RCNN using EdgeBoxes proposals \cite{edgeboxes,fastRcnn} and Deformable Parts Model (DPM) suggested by Felzenszwalb et al. \cite{dpm,objdetdpm}, which is a proven state of the art method in medical domain and tool detection in RAS videos \cite{pote}. The architecture proposed by Girschick et al. \cite{fasterRcnn} scores $90\%$ ($0.9039\%$), takes only $4.21$ hours to train a model with $70k$ iterations and has a mean computation time of $0.59$ seconds per each test frame.  While the improvement with our multimodal approach over FasterRCNN seems small, our architecture has repeatedly and consistently scored higher accuracy for each experiment set. Thus, we believe using a multimodal approach has a stable improvement over the architecture proposed by Girschick et al. \cite{fasterRcnn}, however there might be room for improvement such as pre-training on similar flow data and transferring its weights to initialize our model. This could help our architecture refine its improvement over single modality approach.  In order to test the efficiency of using Region Proposal Network to generate object region proposals, we experiment with an alternative region proposal method. Among the most popular region proposal methods; Selective Search  takes about two seconds per image and generates higher quality proposals while Edge-Boxes takes only 0.2 seconds per image; however, it compromises the quality of the proposals. With initial experimentation, we have observed that the Selective Search did not produce drastically superior results compared to EdgeBoxes on our dataset. We chose to run our experiments with EdgeBoxes with the time consumption concern.  The experimantal results of using EdgeBoxes with Fast RCNN network reach an Average Precision of only $20\%$ while it takes $0.134$ for detection on top of the $2$ seconds to compute the region proposals per frame. We find $40k$ iterations to give better results instead of the $70k$. This approach takes $1.48$ hours to train with $40k$ maximum iterations.  We use a subset of our training images ( each experiment set above $2k$ and on average $2064$ images) to train a Deformable Parts Model (DPM) (voc-release4) with memory concerns. It scores an Average Precision of $76\%$ and up to an average of $83\%$ with bounding box regression. The mean time of object detection by DPM in a test frame is $2.3$ seconds. Each of these experiments are carried with a ten-fold approach, using the same randomized video sets for training and testing. We show sample results in Figure \ref{fig:results}.

\section{Conclusion}

Video understanding in RAS has not yet taken advantage of the recent advances in deep neural networks. In our paper, we propose an end-to-end deep learning approach for fast detection and localization for RAS video understanding. Our architecture applies a Region Proposal Network (RPN), and a multimodal convolutional network for object detection, to jointly predict objectness and localization on a fusion of image and temporal motion cues. We also introduce our dataset ATLAS Dione which provides video data of ten subjects performing six different surgical tasks on the dVSS\textsuperscript{\textregistered} with proper tool annotations. Our experimental results demonstrate that our multimodal architecture is superior to similar methods approaches and also the conventionally used object detection methods in medical domain with an Average Precision (AP) of $91\%$ and a mean computation time of $0.1$ seconds per test frame. With our new architecture that supports multimodality, we improve the results of the architecture proposed by Girschick et al. \cite{fasterRcnn}. Although the improvement is small, our architecture has repeatedly and consistently scored higher accuracy for each experiment set. We believe that making use of the temporal motion cues; optic flow, improves the accuracy by decreasing false positives. Using a fusion of both RGB image and flow modalities make our system more stable and our detections more confident. These findings encourage using additional modalities in detection and localization of tools in RAS videos. Using the architecture proposed by Girschick et al. \cite{fasterRcnn}, it is possible to achieve comparable results to ours in less time for training and testing, this is mainly because our architecture has two different streams of convolutions for RGB input image and flow input image. Although the convolutions are shared for RGB stream and Region Proposal Network, the flow input is convolved in a separate stream before fused, almost doubling the time spent for training and testing, that is excluding the time spent to compute flow images as part for preprocessing. We believe we could further improve our accuracy following the multimodal approach; pre-training on similar flow data and transferring its weights to initialize our model could help us further refine our model. Our results show that using Region Proposal Network jointly with detection network, whether it is the new multimodal architecture we propose or the one proposed by Girschick et al. \cite{fasterRcnn}, dramatically improves the accuracy and reduces the computation time for detection in each frame. We believe our study and dataset will form a benchmark for future studies.


%



\section*{Acknowledgment}

We would like to thank our interns: Lauren Samar, who is a Biomedical Engineering student at Rochester Institute of Technology, and Basel Ahmad, who is a Biomedical Sciences student at SUNY Buffalo, for manual annotation of the data. We would like to acknowledge the ten surgeons working at RPCI who participated this IRB approved study (I 228012). We use the Caffe framework by Jia et al. \cite{caffe} for our experiments and the optical flow estimation code by Brox et al. \cite{flow}, for estimating the temporal motion cues between video frames.

\ifCLASSOPTIONcaptionsoff
  \newpage
\fi

\end{document}